# INTELLIGENT INDOOR MOBILE ROBOT NAVIGATION USING STEREO VISION

Arjun B Krishnan[1] and Jayaram Kollipara[2]

[1] Electronics and Communication Dept., Amrita Vishwa Vidyapeetham, Kerala, India
*abkrishna39@gmail.com*
[2] Electronics and Communication Dept., Amrita Vishwa Vidyapeetham, Kerala, India
*kollipara.jayaram@gmail.com*

## ABSTRACT

*Majority of the existing robot navigation systems, which facilitate the use of laser range finders, sonar sensors or artificial landmarks, has the ability to locate itself in an unknown environment and then build a map of the corresponding environment. Stereo vision, while still being a rapidly developing technique in the field of autonomous mobile robots, are currently less preferable due to its high implementation cost. This paper aims at describing an experimental approach for the building of a stereo vision system that helps the robots to avoid obstacles and navigate through indoor environments and at the same time remaining very much cost effective. This paper discusses the fusion techniques of stereo vision and ultrasound sensors which helps in the successful navigation through different types of complex environments. The data from the sensor enables the robot to create the two dimensional topological map of unknown environments and stereo vision systems models the three dimension model of the same environment.*

## 1. INTRODUCTION

The amount of interest in the field of implementation of robotic systems for tasks like indoor automation, driver-less transportation and the unknown environment exploration have increased exponentially among the community of researchers and engineers. This project addresses the tasks of autonomous navigation and environment exploration using stereo vision based techniques. Other techniques include ultrasound sensors, LIDAR, preloaded maps etc. Out of all these, stereo vision has an edge over other techniques due to its ability to provide three dimensional information about how the environment looks like and decide how obstacles can be avoided for safe navigation through the environment. The currently available stereo cameras are very much expensive and requires special drivers and software to interface with processing platforms. This problem is addressed in this project by making stereo rig using regular webcams thereby making this technique cost-effective.

## 2. RELATED WORKS

Several autonomous mobile robots equipped with stereo vision, were realized in the past few years and deployed both industrially and domestically. They serve humans in various tasks such as tour guidance, food serving, transportation of materials during manufacturing processes, hospital automation and military surveillance. The robots Rhino [1] and Minerva [2] are famous examples of fully operational tour guide robots used in museums which a equipped with stereo vision along with sonar sensors for navigate and map building. The robot Jose [3] uses a Trinocular vision based system that accurately map the environment in all three dimensions.

PR2 [4] is one of the most developed home automation robot which uses a combination of stereo vision and laser range finders for operation

According to [5] there are two essential algorithms for every stereo vision systems: Stereo Calibration algorithm and Stereo Correspondence algorithm. Calibration algorithm is used to extract the parameters of the image sensors and stereo rig, hence has to be executed at least once before using the system for depth calculation. Stereo correspondence algorithm gives the range information by using method of triangulation on matched features. A stereo correspondence algorithm based on global matching is described in [6] uses correspondence search based on block matching. Considering these techniques as a background, an algorithm is designed for this project, which uses horizontal stereo vision system by block matching for obtaining stereo correspondence. Low cost ultrasound sensors and infrared sensors are chosen for overlapping with visual information.

## 3. STEREO VISION BASED OBSTACLE AVOIDANCE

Extraction of 3D position of objects from two or more simultaneous views of a scene is called Stereo vision. Stereo vision systems are dependable and efficient primary sensors for mobile robots and robotic manipulators for extracting the range information from the environment. Stereo vision system can also be used as a tool for all image processing tasks such as colour based segmentation and feature detection, hence serves as the best imaging technique in the field of robotics.

Ideally, the two image sensors in a stereo rig are perfectly aligned along a horizontal or vertical straight line passing through the principle points of both images. Achieving this perfect alignment while mounting the cameras is the main difficulty in realizing custom-made stereo rigs. Moreover, cameras are prone for lens distortions and imaging plane distortions which demand the adoption of Stereo–pair rectification process to remap distorted image projections to undistorted common plane. The obtained rectified images from both the sensors are passed to an algorithm which then searches for matches along each pixel line. The difference in relative positions of an identified feature is called the disparity associated with that feature. Disparity map can be used to understand the depth of objects in the scene with respect to the position of the image sensors. The technique used for mapping the disparities to the real world distances is called triangulation. Figure 1 shows the formation of disparity in stereo image pair using the Pinhole model [7] of two cameras. Robust stereo vision systems are sufficient for segmenting out objects based on their depth, which is an important fact in avoiding collisions during real time navigation. The following sections documents the hardware and software sections of stereo vision system in this project.

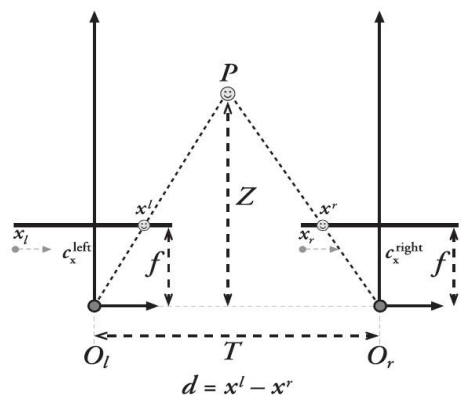

Figure 1. Formation of disparity in a stereo vision system.

### 3.1. The hardware for Stereo Vision System (SVS)

A stereo camera is a type of camera having two or more lenses with separate image sensors for each lens. Stereo vision systems are able to simulate human binocular vision and hence gives the ability to capture three-dimensional images. Two CMOS web cameras, having resolution of 640x480 with USB 2.0 high speed (UVC) interface, are used in this project to make the stereo Rig. An important parameter of a stereo vision system is the baseline length which can be defined as the distance of separation between two cameras, decides the range of depths which can be perceived reliably. The choice of baseline length of a stereo rig is mainly application dependent because a longer baseline length increases both the minimum as well as a maximum bounds of the range while shorter baseline can decrease the bounds [8]. Due to the similarity between the indoor navigation of a robot and a human, the most suitable option for the baseline length is the distance between the human eyes. As a result, a distance of 63 mm is selected as the baseline length for the stereo rig in this project as the mean interpupillary distance of a human is 63.2mm [9]. CAD tool was used to design the mechanical structure of the rig and CNC machine was used to manufacture the designed structure on acrylic sheet. The cameras were fixed with high precision by carefully monitoring collinearity of the obtained left and right images. The stereo rig was covered with opaque film to avoid the exposure to the light from background. The finished stereo rig is shown in Figure 2.

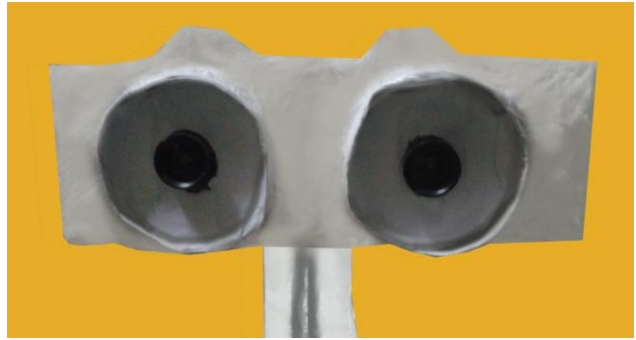

Figure 2. Stereo camera rig made from two webcams.

### 3.2. Algorithms and Software

The algorithms used in this project are developed using OpenCV vision library. OpenCV provides basic as well as advanced functions used in computer vision as an open source package. This library is configured with C++ and used in this project.

The stereo camera will provide simultaneously taken left and right image pairs as an input to the processing unit. Stereo rigs are modelled with Pinhole model and described by Essential matrix E and Fundamental matrix F. Essential matrix relates two cameras with their orientation and Fundamental matrix relates them in pixel coordinates. The initial task for a stereo vision system implementation is to obtain the parameters in these matrices. OpenCV provides predefined functions to find these matrices using RANSAC algorithm [10] and hence calibrate cameras and the rig. Calibration requires a calibration object which is regular in shape and with easily detectable features. The stereo camera calibration algorithm used in this project detects regular chessboard corners from several left and right image pairs taken at different orientations of the chessboard as shown in Figure 3.

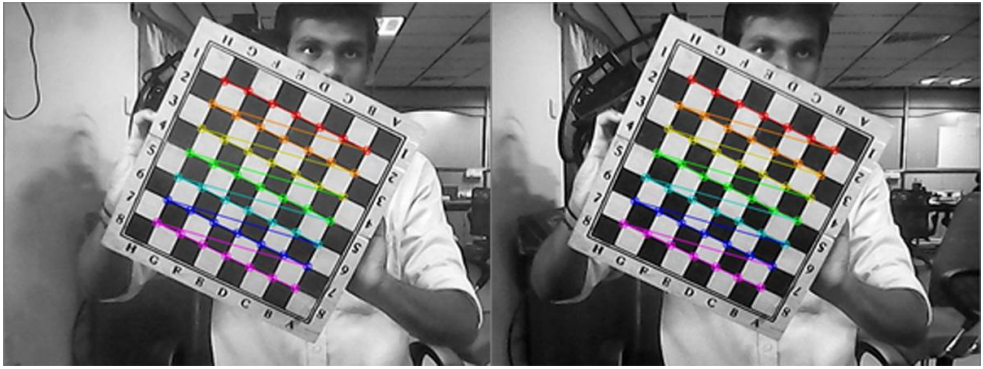

Figure 3. Stereo camera and rig calibration using chessboard as a calibrating object. Detected chessboard corners are marked in simultaneously taken left and right images.

The calibration algorithm computes intrinsic parameters of both the cameras and extrinsic parameters of the stereo rig and stores the fundamental and essential matrixes in a file. This information is used to align image pairs perfectly along the same plane by a process called Stereo Rectification. Rectification enhances both reliability and computational efficiency in depth perception. This is a prime step in the routine if the cameras are misaligned or with an infirm mechanical setup. The custom made stereo setup used in this project showed a negligible misalignment which suggested no requirement of rectification of image pairs for reliable results needed for safe indoor navigation.

The image pair is passed through a block-matching stereo algorithm which works by using small Sum of Absolute Difference (SAD) windows to find matching blocks between the left and right images. This algorithm detects only strongly matching features between two images. Hence the algorithm produces better results for scenes with high texture content and often fails to find correspondence in low textured scenes such as an image of a plane wall. The stereo correspondence algorithm contains three main steps: Pre-filtering of images to normalize their brightness levels and to enhance the texture content, Correspondence search using sliding SAD window of user defined size along horizontal epipolar lines, and post-filtering of detected matches to eliminate bad correspondences.

The speed of the algorithm depends on the size of SAD window and the post-filtering threshold used in the algorithm. Larger SAD windows produce poorer results but elapses less time and vice versa. The choice of window size exhibits a trade-off between quality of the results and algorithm execution time, which leads to the conclusion that this parameter is completely application specific. The window size of 9x9 was selected empirically for the algorithms used in this project. Other parameters associated with the correspondence search algorithm are minimum and maximum disparities of searching. These two values establish the Horopter, the 3D volume that is covered by the search of the stereo algorithm.

The stereo correspondence algorithm generates a greyscale image in which intensity of a pixel is proportional to disparity associated with corresponding pixel location. The obtained disparity values in the image are mapped to real world distances according to the triangulation equation 1.

$$Z = \frac{f \times T}{d} \qquad (1)$$

Where f is the known focal length, T is the distance of separation between cameras, d is the disparity obtained.

Figure 4 shows the disparity map of a scene with four objects at different distances. The low intensity (dark) portions are distant objects whereas high intensity (light) portions are objects which are closer to the camera.

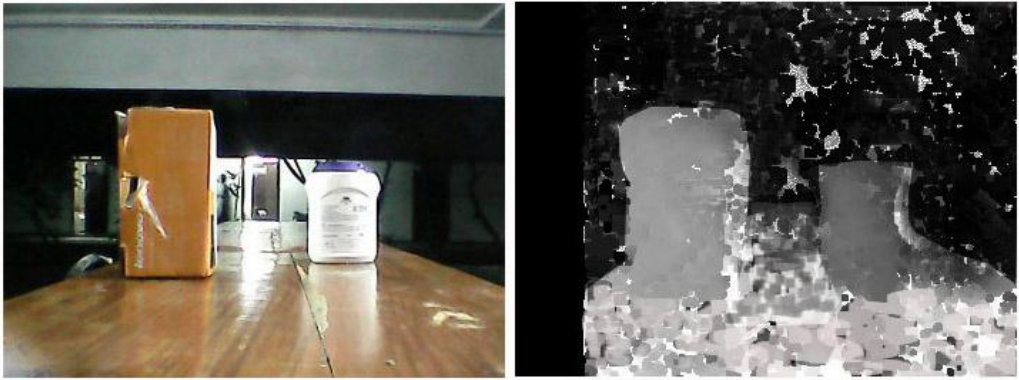

Figure 4: Example of disparity map generated using stereo vision system. Low intensity areas correspond to farther objects and high intensity portions are nearer objects.

### 3.3. Depth based Image segmentation for obstacle avoidance

The disparity maps generated by above mentioned algorithm plays a vital role in obstacle avoidance during navigation. The segmentation based on the intensity levels is same as segmentation based on depth. The disparity images are dilated using 3x3 rectangular mask to fill small holes present in disparity. A segmentation algorithm is used to detect near objects which isolates regions which are having high intensity range and searches for connected areas that can form blobs within the segmented regions. The intensity range for segmentation is determined experimentally such a way that all the obstacles in 20 cm to 40 cm are detected. The contours of these blobs are detected and bounding box coordinates for each blobs are calculated. The centres of the bounding boxes as well as the bounding boxes are marked on the image. The input image from left camera is divided into two halves to classify the position of the detected object to left or right. The centre of the contour is tracked and if it is found out to be in the left half of the image, algorithm takes a decision to turn the robot to the right side and vice versa. If no obstacles are found in the search region robot will continue in its motion along the forward path. In case of multiple object occurrences in both halves, robot is instructed to take a 90 degree turn and continue the operation. Figure 5 shows the disparity map of several obstacle conditions and the corresponding decisions taken by the processing unit in each case.

Instruction from processing unit is communicated with robot's embedded system through USART communication. Instruction to move forward will evoke the PID algorithm implemented and robot follows exact straight line path unless the presence of an obstacle is detected by the vision system. Our algorithm elapses 200 ms for a single decision making. Dynamic obstacles such as moving humans may not be properly detected by the stereo vision. But this issue is handled by giving high priority for ultrasound sensors and the robot is able to stop instantly. Obstacle detection from ultrasound sensors interrupts the stereo vision algorithm and directly instructs the robot to stop the embedded system level itself. After stopping, control is immediately handed over to the processing unit for deciding distance and shape of the obstacles.

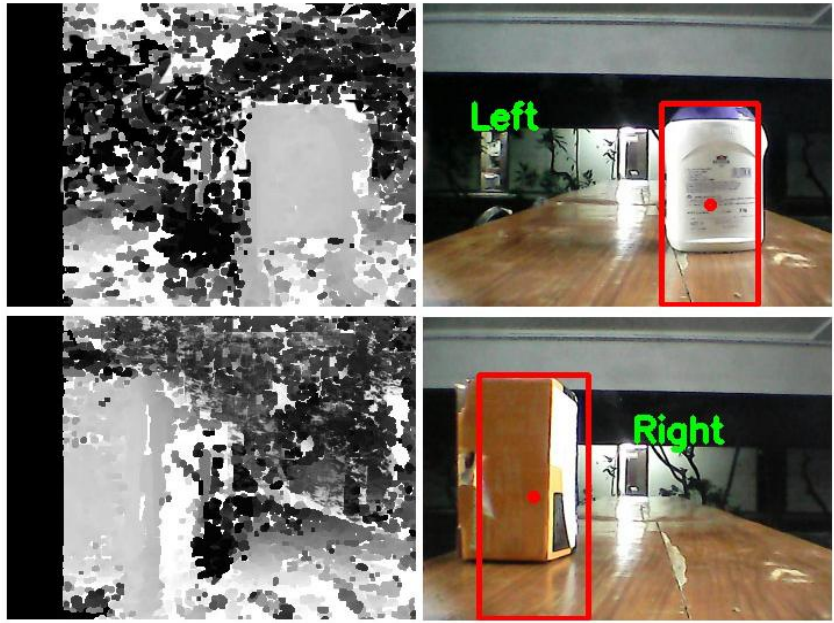

Figure 5. Disparity map of several obstacle conditions in an indoor environment (left). Detected obstacles in the specified distance range and corresponding decisions taken are shown (right)

## 4. 3D RECONSTRUCTION

Three Dimensional reconstruction is the process of generating the real world model of the scene observed by multiple views. Generated disparity maps from each scenes can be converted into corresponding point clouds with real world X, Y and Z coordinates. The process of reconstruction of 3D points requires certain parameters obtained from the calibration of Stereo rig. An entity called Re-projection matrix is formed from the intrinsic and extrinsic parameters and it denotes the relation between real world coordinates and pixel coordinates. Re-projection matrix is formed during the calibration steps. The entries of re-projection matrix is shown in Figure 6.

$$Q = \begin{bmatrix} 1 & 0 & 0 & -c_x \\ 0 & 1 & 0 & -c_y \\ 0 & 0 & 0 & f \\ 0 & 0 & -1/T_x & (c_x - c'_x)/T_x \end{bmatrix}$$

Figure 6. Re-projection matrix of a Stereo Rig

($c_x$, $c_y$) – is the principal point of the camera. The point at which the image plane coincides exactly with the middle point of the lens.
f – Focal length of the camera, as the cameras in the stereo rig are set to same focal length thus the Re-projection matrix has a single focal length parameter.
$T_x$ – Translation coefficient in x –direction.

The Re-projection matrix thus generated converts a disparity map into a 3D point cloud by using the matrix computation shown in equation 2.

$$Q \begin{bmatrix} x \\ y \\ d \\ 1 \end{bmatrix} = \begin{bmatrix} X \\ Y \\ Z \\ W \end{bmatrix} \qquad (2)$$

Where *x* and *y* are the coordinates of a pixel in the left image, *d* is the corresponding disparity associated with that pixel and Q is the re-projection matrix. The real world coordinates can be computed by dividing X, Y and Z by W present in the output matrix.

The calculated 3d point clouds and their corresponding RGB pixel values are stored in the memory file in a text file along with the odometric references at each instance of point cloud generation. The stored point cloud is retrieved and filtered using Point Cloud Library (PCL) integrated with C++. Point clouds groups having a cluster size above a particular threshold level are only used in 3D reconstruction and thereby removing noisy point clusters. Point clouds beyond the threshold distance are also removed since the error of projection increases with increasing real world distance.3D reconstructions are generated and stored according to the alignment of the robot. The complete 3D mapping of an environment is obtained by the overlapped re-projection of continuous scenes. This 3D map can be used to plan the path if a destination point is provided the robot .The visualised 3D reconstruction examples are shown in Figure 7.

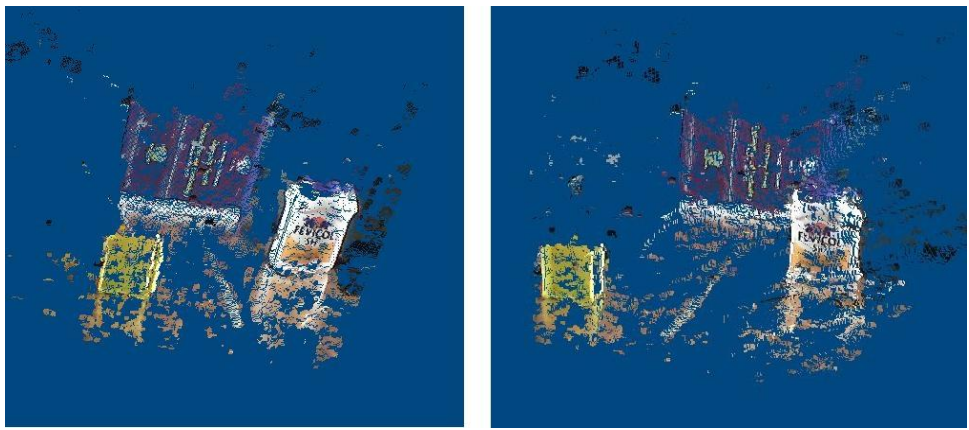

Figure 7. 3D Reconstructions of filtered Point clouds

## 5. EXPERIMENTAL ROBOTIC PLATFORM

The experimental mobile robotic platform used in this project is a six wheeled differential drive rover which is able to carry a portable personal computer. There are three ultrasound sensors attached to the front of the robot. Vertical depth information of the operating surface is monitored by two infrared range finders thereby avoiding falling from an elevated surfaces. A three axis digital compass module is used to find the direction of robot's movement. High torque geared motors of 45 RPM are used to power the four wheels which gives the robot a velocity of 20cm/sec. Optical encoders are attached other two free rotating wheels for the keeping track of the distance travelled. The optical encoder generates 400 pulses per revolution and hence gives a resolution of 0.90 degrees. The core elements of the embedded system of this robot are two 8 bit ATmega328 Microcontroller based Arduino boards. One Arduino collects information from optical wheel encoders based on interrupt based counting technique and the other collects data from all other sensors and controls the motion of the motors through a motor driver. Reliable odometric feedback are provided to the control system through the heading from compass and distance data from wheel encoders. A PID algorithm has been implemented to keep the robot along the exact path planned by the vision system in an obstacle free region.

The feedback for PID algorithm is the direction of heading obtained from digital compass. Arduino boards transfer data from the sensors to the on board PC for storage and receives decisions from vision system implemented in on-board PC over a USB to USART serial converter module link.

## 6. RESULTS

One of the rapidly developing but least pondered research area of Stereo vision based SLAM architecture has been dealt in this project. We have been able to successfully introduce a cost effective prototype of the stereo camera and robotic platform. Outputs comparable with commercially available alternatives can be provided from the Stereo Vision System. The stereo matching program can process five frames per second in a 1.6 GHz Intel atom processor board equipped with 2 GB RAM. This is an adequate performance for safe indoor navigation for slow moving robots. An almost error–proof navigation for robot in indoor environment is ensured with the process of overlapping of vision perception with other information from sensors. An accurate 2D mapping of the environment based on the ultrasound data and 3D mapping using stereo vision has been implemented. For a sample data collected from a test run timed four minutes, 3D reconstruction elapses 25 to 80 ms per frame whereas 2D mapping requires less than 50 ms time. A sense of intelligence is given to the robot through the detection of objects using vision just as in the case of the human vision. It has also been proved that for the successful completion of tasks identified during the proposal of the project, the choice of mechanical parameters of stereo rig, range of the horopter, stereo correspondence algorithm parameters and filter parameters made in this project are very much sufficient.

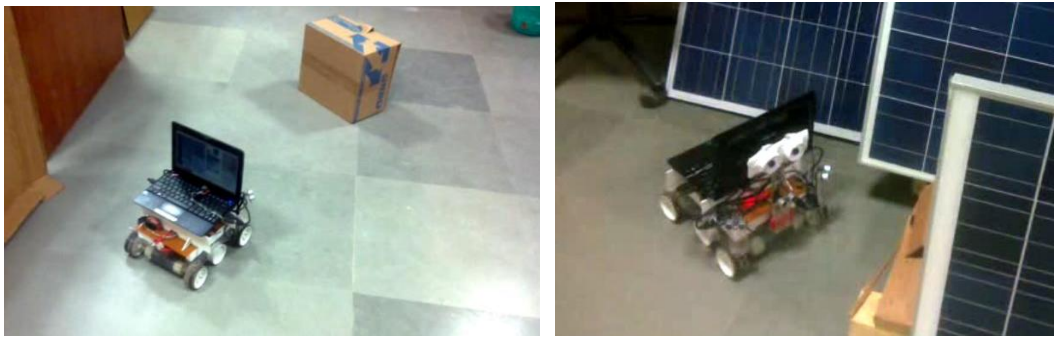

Figure 8. Robot operates in cluttered indoor environment

## 7. DISCUSSION AND FUTURE WORK

This paper outlines the implementation of a cost-effective stereo vision system for a slowly moving robot in an indoor environment. The detailed descriptions of algorithms used for stereo vision, obstacle avoidance, navigation and three dimensional map reconstruction are included in this paper. The robot described in this paper is able to navigate through a completely unknown environment without any manual control. The robot can be deployed to explore an unknown environment such as collapsed buildings and inaccessible environments for soldiers during war. Vision based navigation allows robot to actively interact with the environment. Even though vision based navigation systems are having certain drawbacks when compared with other techniques. Stereo vision fails when it is being subjected to surfaces with less textures and features, such as single colour walls and glass surfaces. The illumination level of environment is another factor which considerably affects the performance of stereo vision. The choice of processing platform is crucial in the case of processor intense algorithms used in disparity map generation. Point clouds generated are huge amount of data which has to be properly handled and saved for better performances.

The future works related to this project are developing of a stereo camera which has reliable disparity range over longer distance, implementing the stereo vision algorithm in a dedicated processor board and further development of the robot for outdoor navigation with the aid of Global Positioning System.

## REFERENCES

[1]  J. Buhmann, W. Burgard, A.B. Cremers, D. Fox, T. Hofmann, F. Schneider, J. Strikos, and S. Thrun, (1995) "The mobile robot Rhino," *AI Magazine*, Vol. 16, No. 1.

[2]  S. Thrun, M. Bennewitz, W. Burgard, A.B. Cremers, F. Dellaert, D. Fox, D. Hähnel, C. Rosenberg, N. Roy, J. Schulte and D. Schulz, (1999) "MINERVA: A second generation mobile tour-guide robot," in *Proc. IEEE International Conference on Robotics and Automation (ICRA)*, vol.3, No., pp.1999.

[3]  Don Murray, and Jim Little, (2000) "Using real-time stereo vision for mobile robot navigation," *Autonomous Robots*, Vol. 8, No. 2, pp.161-171.

[4]  Pitzer, B., Osentoski, S., Jay, G., Crick, C., and Jenkins, O.C., (2012) "PR2 Remote Lab: An environment for remote development and experimentation," *Robotics and Automation (ICRA)*, vol., no., pp.3200 – 3205.

[5]  Kumar S., (2009) "Binocular Stereo Vision Based Obstacle Avoidance Algorithm for Autonomous Mobile Robots," Advance Computing Conference, *IACC 2009. IEEE International*, vol., no., pp.254-259.

[6]  H. Tao, H. Sawhney, and R. Kumar. (2001) "A global matching framework for stereo computation," In *Proc. International Conference on Computer Vision*, Vol. 1.

[7]  Z. Zhang, G. Medioni and S.B. Kang, (2004) "Camera Calibration", *Emerging Topics in Computer Vision*, Prentice Hall Professional Technical Reference, Ch. 2, pp.4-43.

[8]  M. O kutomi and T . K anade, (1993) "A multiple-baseline stereo," *IEEE Transactions on Pattern Analysys and Machine Intelligence*, Vol. 15, No. 4, pp.353-363.

[9]  Dodgson, N. A, (2004) "Variation and extrema of human interpupillary distance," In A. J. Woods, J. O. Merritt, S. A. Benton and M. T. Bolas (eds.), *Proceedings of SPIE: Stereoscopic Displays and Virtual Reality Systems XI,* Vol. 5291, pp.36–46.

[10] M.A. Fischler and R.C. Bolles, (1981) "Random sample consensus: a paradigm for model fitting with application to image analysis and automated cartography". *Communication of ACM*, Vol. 24, No. 6, pp.381–95.

[11] G. Bradski and A. Kaehler, (2008) "Learning OpenCV: Computer Vision with the OpenCV Library," O'Reilly Media, Inc.

[12] Murray, D. and Jennings, C., "Stereo vision based mapping and navigation for mobile robots," in *Proc. 1997 IEEE International Conference on Robotics and Automation*, Vol. 2, pp.1694-1699.

**Authors**

**Arjun B Krishnan** received Bachelor of Technology degree in Electronics and Communication Engineering from Amrita Vishwa Vidyapeetham, Kollam, India in 2014. Currently, he is working as a researcher in Mechatronics and Intelligent Systems Research Laboratory under Mechanical Dept. of Amrita Vishwa Vidyapeetham. His research interests include Autonomous mobile robotics, Computer vision and Machine learning.

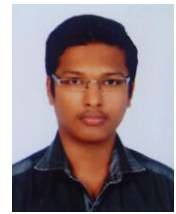

**Jayaram Kollipara** received Bachelor of Technology degree in Electronics and Communication Engineering from Amrita Vishwa Vidyapeetham, Kollam, India in 2014. He joined as a Program Analyst in Cognizant Technology Solutions, India. His research interests are Image and Signal processing, Pattern recognition and Artificial intelligence.

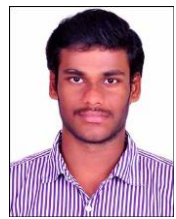